\documentclass[11pt]{article}

\PassOptionsToPackage{table}{xcolor}
\usepackage[preprint]{acl}

\usepackage{times}
\usepackage{latexsym}
 
\usepackage[T1]{fontenc}

\usepackage[utf8]{inputenc}

\usepackage{microtype}
\usepackage{inconsolata}
\usepackage{graphicx}
\usepackage{amsmath}
      
\usepackage{caption}
\usepackage{subcaption}
\usepackage{booktabs}
\usepackage{array}
\usepackage{diagbox}
\usepackage{float}
\usepackage{comment}
\usepackage{cleveref}

\title{Guess the Unified Model: How Much Can We Recover from\\ Generated Images?}

\author{
  Jasin Cekinmez\textsuperscript{*} \quad Ryo Mitsuhashi\textsuperscript{*} \quad Addison J. Wu\textsuperscript{*} \quad Yida Yin \\
  Princeton University \\
  \texttt{\{jasincekinmez, rm4411, addisonwu\}@princeton.edu}
}

\newcommand\blfootnote[1]{%
  \begingroup
  \renewcommand\thefootnote{}\footnote{#1}%
  \addtocounter{footnote}{-1}%
  \endgroup
}

\begin{document}
\maketitle
\blfootnote{\textsuperscript{*}Equal contribution. Author order selected randomly.}

\begin{abstract}
With unified model-generated images now widespread online, attributing their model of origin offers a path toward transparency and deeper insight into the characteristic behaviors of individual models. Prior work has explored provenance in LLM-generated text, diffusion model images, and datasets, but the separability of unified model-generated images remains an underexplored area. We address this gap by examining separability across corruption, domains, and prompt languages using images generated by seven unified models. We show that model attribution is highly feasible as our model achieves near-perfect accuracy with around 20K images per model. Corruptions and structural perturbations have only a modest effect on attribution performance, and cross-domain generalization reveals that semantic content contributes to separability but is not the dominant signal. Finally, we observe that for most models, prompt language attribution is around chance levels, suggesting minimal language-specific visual signatures. These findings highlight consistent model-specific visual characteristics in unified models outputs and open new directions for tracing and auditing generative image pipelines.
\end{abstract}

\section{Introduction}
\begin{figure*}[t]
    \centering
    \begin{subfigure}[b]{0.13\textwidth}
        \centering
        \includegraphics[width=\textwidth]{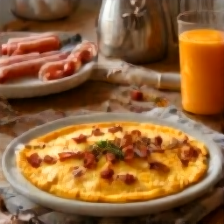}
        \caption{BAGEL}
    \end{subfigure}
    \hfill
    \begin{subfigure}[b]{0.13\textwidth}
        \centering
        \includegraphics[width=\textwidth]{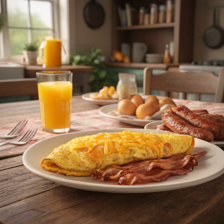}
        \caption{Emu3.5}
    \end{subfigure}
    \hfill
    \begin{subfigure}[b]{0.13\textwidth}
        \centering
        \includegraphics[width=\textwidth]{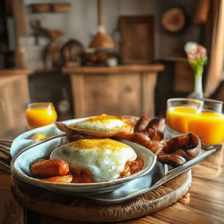}
        \caption{Janus}
    \end{subfigure}
    \hfill
    \begin{subfigure}[b]{0.13\textwidth}
        \centering
        \includegraphics[width=\textwidth]{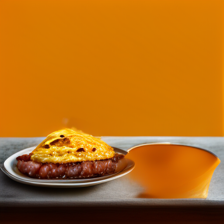}
        \caption{MMaDA}
    \end{subfigure}
    \hfill
    \begin{subfigure}[b]{0.13\textwidth}
        \centering
        \includegraphics[width=\textwidth]{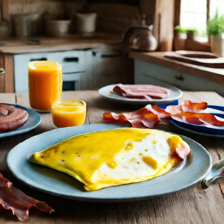}
        \caption{Show-o2}
    \end{subfigure}
    \hfill
    \begin{subfigure}[b]{0.13\textwidth}
        \centering
        \includegraphics[width=\textwidth]{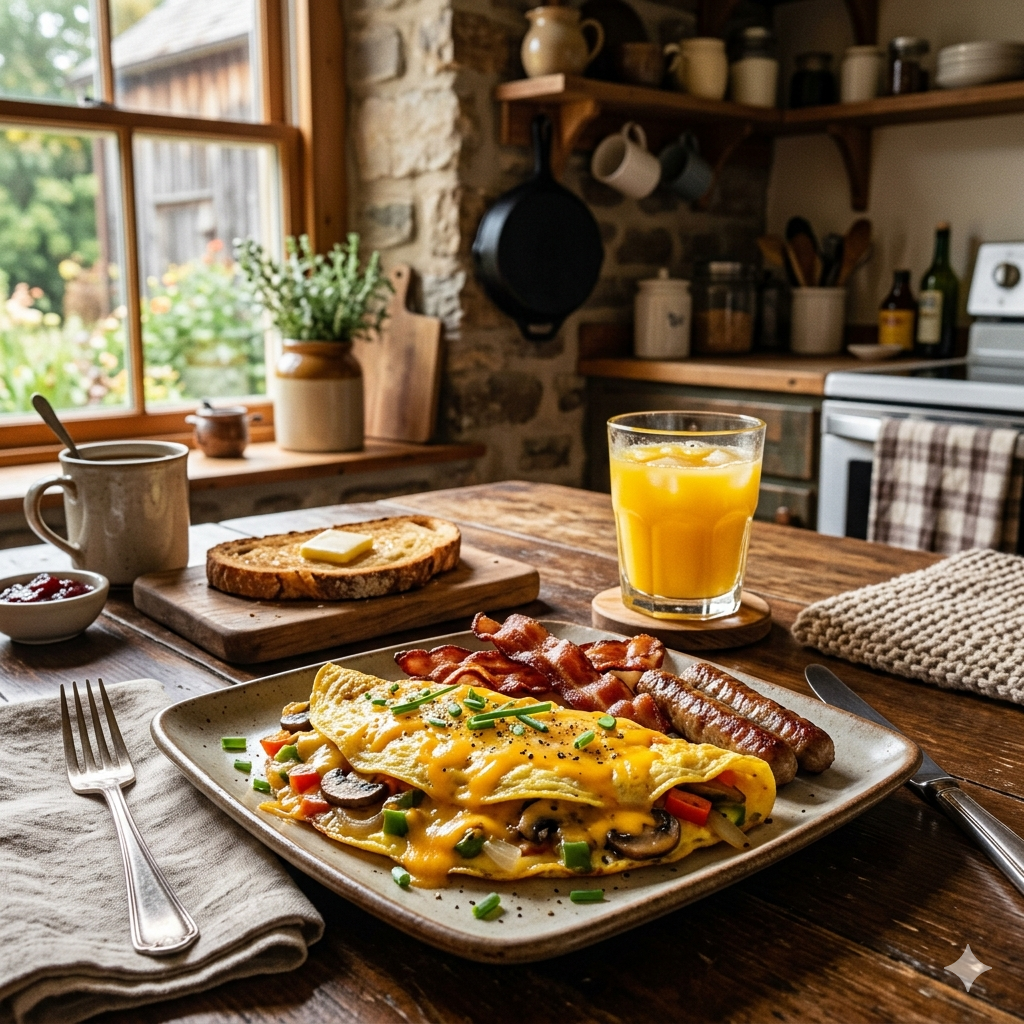}
        \caption{Gemini}
    \end{subfigure}
    \hfill
    \begin{subfigure}[b]{0.13\textwidth}
        \centering
        \includegraphics[width=\textwidth]{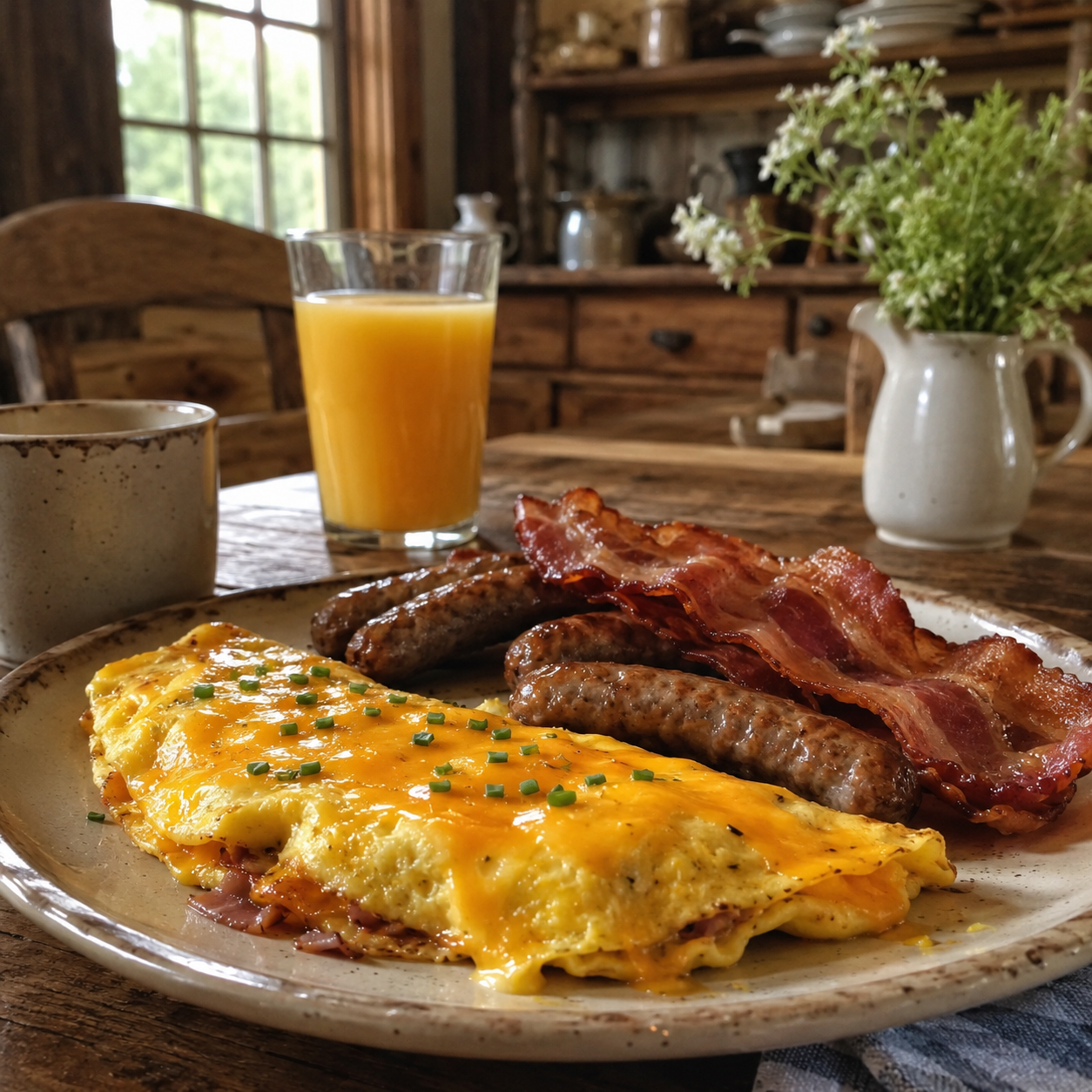}
        \caption{GPT}
    \end{subfigure}
    \caption{Images generated by unified models with the prompt ``american breakfast, photography, rustic farm house kitchen and dining room. omelette cheesey photography 4k, sausage links and bacon breakfast meat. served dishes. orange juice background''}
    \label{fig:img-comparison}
\end{figure*}
With the rapid advancement of large language models (LLMs), recent years have seen these systems expand beyond text into images, audio, and video, giving rise to unified multimodal models capable of jointly understanding and generating diverse forms of data. Models such as the GPT~\citep{openai2024gpt4technicalreport}, Gemini~\citep{geminiteam2025geminifamilyhighlycapable}, and Chameleon~\citep{team2024chameleon} families now produce high-quality images that are increasingly difficult to distinguish from real content and are widely deployed across online platforms. As a result, model-generated images have become pervasive in real-world settings, raising urgent questions about transparency, accountability, and the ability to trace synthetic content back to its source.

Understanding the origin of synthetic images is essential for characterizing model-specific behaviors and failure modes, as well as for responding to misuse such as misinformation or deceptive content. Prior work has explored provenance and attribution, including dataset attribution for analyzing visual bias~\citep{zeng2024understandingbiaslargescalevisual, liu2025decadesbattledatasetbias}, and attribution of LLM-generated text~\citep{sun2025idiosyncrasieslargelanguagemodels} and diffusion model-generated images~\citep{xu2025detectingoriginattributiontexttoimage, bongini2025training}. However, images generated by unified models differ fundamentally from those produced by diffusion pipelines~\citep{ho2020denoising, rombach2022high}, as they are conditioned on language, vision, and internal cross-modal representations. Despite their deployment, the visual separability of images generated by frontier unified models remains underexplored.

We find that a classifier based on five open-source unified models achieves high accuracy on this task even with limited training images, indicating clear idiosyncrasies in the models. Notably, with 500 training images per model, the classifier achieves 59.8\% accuracy on a held-out test set, compared to the random chance of 20\%. With 25K images per model, the accuracy reaches 99.9\%.

We analyze what factors influence classification accuracy by applying corruptions and structural perturbations. We find that the images are separable in high-level features, spatial structure, object-level structure, and color distribution. We also observe that image attribution is not driven by the semantic content of the images by conducting out-of-distribution experiments.

Lastly, we investigate whether the prompt language affects the distribution of the generated images and find that the distribution of generated images does not depend on the prompt language for many state-of-the-art models. Specifically, we conduct image attribution tasks across five languages for five open-source and two closed-source unified models and find that the classification accuracy stays at the chance-level 20\% for two of the open-source models and both of the closed-source models.
\section{Related Work}

\paragraph{Dataset Classification.} Prior work on dataset classification has shown that visual data sources often leave strong, identifiable signatures. \citet{torralba2011unbiased} demonstrated that image datasets from the 2010s could be reliably distinguished by a classifier, revealing substantial dataset-specific biases. More recently, \citet{liu2025decadesbattledatasetbias} scaled this analysis to larger and more diverse datasets and found that, despite increased diversity, datasets remain highly separable, indicating that systematic biases persist. Similar conclusions have been drawn in large-scale visual settings~\citep{zeng2024understandingbiaslargescalevisual}. Unlike dataset classification, which captures biases arising from data collection and curation, unified model classification focuses on biases introduced during model creation and training. This helps us better understand how design and training choices shape the biases seen in generated images.

\paragraph{Generative Model Landscape.} Generative image modeling has progressed from GAN-based methods~\citep{goodfellow2014generative} to diffusion models~\citep{ho2020denoising, rombach2022high}, and most recently to unified architectures that jointly handle language and vision. Joint vision-language pretraining~\citep{radford2021learning} and large-scale paired datasets~\citep{schuhmann2022laion} enabled this transition. Text-conditional generators such as DALL-E~2~\citep{ramesh2022hierarchical} represent the diffusion paradigm, while newer unified architectures—including Chameleon~\citep{team2024chameleon}, Transfusion~\citep{zhou2024transfusion}, Show-o~\citep{xie2024show}, and Janus~\citep{wu2024janus}—integrate image understanding and generation in a single model. Unlike purely diffusion-based generators, these unified models condition on shared cross-modal representations, which may give rise to distinct generative signatures that are the focus of our study.

\paragraph{Attribution of Machine-Generated Content.} Prior work demonstrates that generated content can often be attributed to the specific model that produced it. In the image domain, GAN-generated images were found to carry detectable model fingerprints~\citep{marra2019gans, gragnaniello2021gan, wang2020cnn}, and this has since extended to diffusion-generated images~\citep{corvi2023detection, bongini2025training, xu2025detectingoriginattributiontexttoimage}. In the text domain, classifier-based approaches~\citep{sun2025idiosyncrasieslargelanguagemodels}, watermarking~\citep{kirchenbauer2023watermark}, and zero-shot statistical detection~\citep{mitchell2023detectgpt} have all been used to attribute or detect LLM-generated text. A systematic study of image attribution in unified models is warranted, as it remains underexplored and is central to understanding whether generated images can be reliably traced back to their source model.

\paragraph{Prompt Language Effects.} Beyond model identity, recent work raises questions about what linguistic and cultural information is preserved or homogenized during image generation. \citet{shi2025culture} show that text-to-image models often produce culturally Westernized imagery even when prompted in non-Western languages, suggesting a collapse of culturally specific conditioning in the generation process. Other studies demonstrate that properties of the prompt language, such as whether it is grammatically gendered and the gender stereotypes embedded in that language or culture, can significantly influence generated images, particularly with respect to gender presentation and social roles~\citep{friedrich2024multilingual, saeed2025beyond}. More broadly, text-to-image models have been shown to amplify demographic stereotypes~\citep{bianchi2023easily} and exhibit social biases in visual reasoning~\citep{cho2023dall}. These findings present a tension, while some cultural signals appear to be erased, others are strongly preserved and propagated through generations. This raises a broader question of whether prompt language leaves identifiable traces in the output of unified models, and whether such traces could be leveraged to infer properties of the input language. Understanding whether and how language-specific information remains separable in these models has important implications for robustness, bias, and security-sensitive applications.

\section{Methodology}

\subsection{Classification Task}
Following the dataset classification formulation introduced in \citet{liu2025decadesbattledatasetbias}, we treat attribution as a multi-class classification problem over data sources. \citet{liu2025decadesbattledatasetbias} evaluates separability between datasets, serving as a diagnostic for systematic biases or statistically meaningful differences in their underlying distributions.

We adapt this formulation to unified model attribution by treating each of the $N$ unified models as a distinct class. Given an image, the task is to predict which unified model generated the image, resulting in an $N$-way classification problem. This reframing allows us to quantify attribution fidelity by measuring how well model-specific signals are preserved and separable across unified models.

\subsection{Classifier and Image Generation Setup}
We use ConvNeXT~\citep{liu2022convnet} as our classifier since it offers strong accuracy and computational efficiency for this task. We prefer ConvNeXT over transformer-based vision architectures~\citep{dosovitskiy2021image}, given our relatively small training sets. We use the ConvNeXT-Tiny model because we observed no improvement from larger ConvNeXT variants in preliminary experiments.

We generate images using the same prompts across all unified models and languages, so that the classifier learns model-specific visual cues rather than differences in subject matter. We then train a ConvNeXT classifier from scratch on the resulting training set and evaluate performance on a held-out test set. In all the experiments, we train a classifier until 200 epochs and measure the accuracy on a held-out test set at the last iteration. We used a learning rate of 1e-3 and warmup epochs of 2. Batch size ranges from 32 to 256, depending on the number of training data. All the images from unified models are generated with a 1:1 aspect ratio with default resolution and resized into $224 \times 224$ before given to the classifier.

\subsection{Overview of Experiments}
We conduct five experiments. The first three use images generated by open-source models, while the latter two include images generated by both open- and closed-source models. The open-source models are BAGEL~\citep{deng2025emerging}, MMaDA~\citep{yang2025mmada}, Emu-3.5~\citep{cui2025emu3}, DeepSeek Janus-Pro-7B~\citep{chen2025janus}, and Show-o2~\citep{xie2025show}. The closed-source models are Gemini 2.5 Flash Image (Nano Banana) and GPT-Image-1. See \Cref{fig:img-comparison} for example images from each model.

The first experiment explores the effects of scaling, where we vary the number of training examples to measure how separability changes with size. The second experiment examines corruption, where we apply corruptions to the images prior to training to test how separability depends on low-level features. The third experiment examines how the structural bias influences separability by applying structural perturbations to the images. The fourth experiment studies Out-of-Distribution (OOD) generalization, where we train on a specific domain and evaluate on all domains. Lastly, the language experiment tests whether we can identify the prompt language conditioned on knowing the unified model, in order to determine whether language provides any visual cues.

\section{Model Attribution \& Scaling Performance}

\subsection{Classifier Scaling Trends}
\begin{figure}[t]
    \includegraphics[width=\columnwidth]{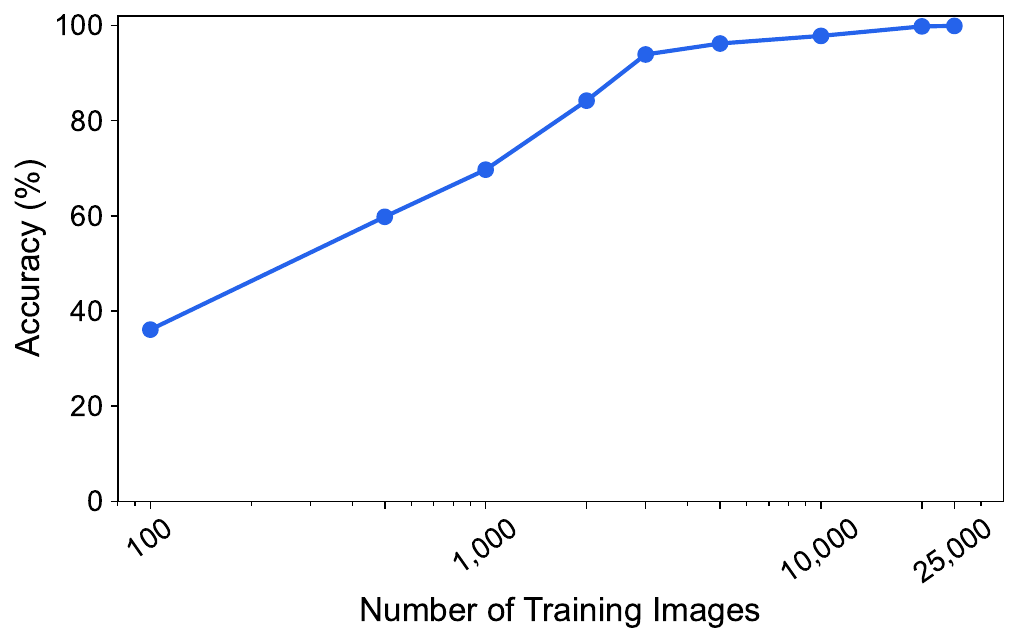}
    \caption{Attribution accuracy scales rapidly, surpassing 90\% at 3K images/model and reaching 99.9\% at 25K.}
    \label{fig:scaling}
\end{figure}

We utilize the MJHQ-30K dataset~\citep{li2024playground} to generate images for the 5 open-source models for this experiment since it provides a large collection of complex and descriptive prompts, allowing us to examine separability under controlled and realistic prompting conditions. We trained a 5-way classifier (BAGEL, Emu3.5, Janus, MMaDA, Show-o2) with different numbers of training images, ranging from 100 images per model to 25K images per model. For every training run, we used a held-out test set containing 5K images per model. With 100 images per model, the classifier achieved 36\% accuracy over the 5 models, which surpasses a random guess of 20\%. With 3K training images per model, the classifier achieved an accuracy of \textgreater 90\%, and with 25K training images per model, the accuracy reached 99.9\% as shown in \Cref{fig:scaling}.

While discerning models from images is not challenging for the classifier, some models are easier to discern than others. \Cref{tab:combined_stats} reveals asymmetries in how different models are identified. Janus and Show-o2 have the highest recall rates (57.3\% and 61.2\%) when the number of training images is 100 per model, indicating that their outputs are more separable than the remaining models. In contrast, Emu achieves a relatively high precision rate but one of the lowest recall rates, showing that although the classifier does not identify Emu correctly very often, when it does make an Emu prediction, it is usually correct. This suggests that Emu's outputs may contain some distinct visual characteristics, but these cues are not consistently present. Overall, this result highlights that separability is not uniform across models.
Qualitatively, we found that the images generated by Emu are the most consistent and rarely disfigured, and the images by MMaDA often have emptier backgrounds than others, as shown in \Cref{fig:img-comparison}, which could explain the high precision for Emu and MMaDA.

\begin{table*}[ht]
    \centering

    \begin{subtable}[b]{0.48\textwidth}
        \centering
        \resizebox{\columnwidth}{!}{
\begin{tabular}{lw{c}{1.2cm}w{c}{1.2cm}w{c}{1.2cm}w{c}{1.2cm}w{c}{1.2cm}}
\diagbox{\textbf{True}}{\textbf{Pred}} & \rotatebox{45}{BAGEL} & \rotatebox{45}{Emu} & \rotatebox{45}{Janus} & \rotatebox{45}{MMaDA} & \rotatebox{45}{Show-o2} \\
\text{BAGEL} & \cellcolor[rgb]{0.270,0.015,0.341}\parbox[c][1.2cm][c]{1.2cm}{\centering \textcolor{white}{1.6}} & \cellcolor[rgb]{0.282,0.090,0.412}\parbox[c][1.2cm][c]{1.2cm}{\centering \textcolor{white}{4.8}} & \cellcolor[rgb]{0.132,0.552,0.553}\parbox[c][1.2cm][c]{1.2cm}{\centering \textcolor{white}{30.4}} & \cellcolor[rgb]{0.214,0.356,0.551}\parbox[c][1.2cm][c]{1.2cm}{\centering \textcolor{white}{18.1}} & \cellcolor[rgb]{0.328,0.774,0.407}\parbox[c][1.2cm][c]{1.2cm}{\centering \textcolor{black}{45.1}} \\
\text{Emu} & \cellcolor[rgb]{0.269,0.010,0.335}\parbox[c][1.2cm][c]{1.2cm}{\centering \textcolor{white}{1.4}} & \cellcolor[rgb]{0.268,0.224,0.512}\parbox[c][1.2cm][c]{1.2cm}{\centering \textcolor{white}{11.0}} & \cellcolor[rgb]{0.165,0.467,0.558}\parbox[c][1.2cm][c]{1.2cm}{\centering \textcolor{white}{25.0}} & \cellcolor[rgb]{0.153,0.497,0.558}\parbox[c][1.2cm][c]{1.2cm}{\centering \textcolor{white}{26.7}} & \cellcolor[rgb]{0.125,0.640,0.527}\parbox[c][1.2cm][c]{1.2cm}{\centering \textcolor{black}{35.9}} \\
\text{Janus} & \cellcolor[rgb]{0.269,0.010,0.335}\parbox[c][1.2cm][c]{1.2cm}{\centering \textcolor{white}{1.4}} & \cellcolor[rgb]{0.275,0.038,0.365}\parbox[c][1.2cm][c]{1.2cm}{\centering \textcolor{white}{2.6}} & \cellcolor[rgb]{0.835,0.886,0.103}\parbox[c][1.2cm][c]{1.2cm}{\centering \textcolor{black}{57.3}} & \cellcolor[rgb]{0.199,0.388,0.555}\parbox[c][1.2cm][c]{1.2cm}{\centering \textcolor{white}{20.0}} & \cellcolor[rgb]{0.211,0.364,0.552}\parbox[c][1.2cm][c]{1.2cm}{\centering \textcolor{white}{18.7}} \\
\text{MMaDA} & \cellcolor[rgb]{0.267,0.005,0.329}\parbox[c][1.2cm][c]{1.2cm}{\centering \textcolor{white}{1.1}} & \cellcolor[rgb]{0.283,0.111,0.432}\parbox[c][1.2cm][c]{1.2cm}{\centering \textcolor{white}{5.7}} & \cellcolor[rgb]{0.142,0.526,0.556}\parbox[c][1.2cm][c]{1.2cm}{\centering \textcolor{white}{28.8}} & \cellcolor[rgb]{0.395,0.797,0.368}\parbox[c][1.2cm][c]{1.2cm}{\centering \textcolor{black}{47.0}} & \cellcolor[rgb]{0.220,0.343,0.549}\parbox[c][1.2cm][c]{1.2cm}{\centering \textcolor{white}{17.4}} \\
\text{Show-o2} & \cellcolor[rgb]{0.271,0.020,0.347}\parbox[c][1.2cm][c]{1.2cm}{\centering \textcolor{white}{1.9}} & \cellcolor[rgb]{0.280,0.068,0.392}\parbox[c][1.2cm][c]{1.2cm}{\centering \textcolor{white}{3.8}} & \cellcolor[rgb]{0.176,0.441,0.558}\parbox[c][1.2cm][c]{1.2cm}{\centering \textcolor{white}{23.4}} & \cellcolor[rgb]{0.275,0.195,0.496}\parbox[c][1.2cm][c]{1.2cm}{\centering \textcolor{white}{9.7}} & \cellcolor[rgb]{0.993,0.906,0.144}\parbox[c][1.2cm][c]{1.2cm}{\centering \textcolor{black}{61.2}} \\
\end{tabular}
}
        \caption{Recall (Row-normalized)}
        \label{tab:recall}
    \end{subtable}
    \hfill
    \begin{subtable}[b]{0.48\textwidth}
        \centering
        \resizebox{\columnwidth}{!}{
\begin{tabular}{lw{c}{1.2cm}w{c}{1.2cm}w{c}{1.2cm}w{c}{1.2cm}w{c}{1.2cm}}
\diagbox{\textbf{True}}{\textbf{Pred}} & \rotatebox{45}{BAGEL} & \rotatebox{45}{Emu} & \rotatebox{45}{Janus} & \rotatebox{45}{MMaDA} & \rotatebox{45}{Show-o2} \\
\text{BAGEL} & \cellcolor[rgb]{0.173,0.449,0.558}\parbox[c][1.2cm][c]{1.2cm}{\centering \textcolor{white}{19.7}} & \cellcolor[rgb]{0.209,0.368,0.553}\parbox[c][1.2cm][c]{1.2cm}{\centering \textcolor{white}{17.1}} & \cellcolor[rgb]{0.184,0.422,0.557}\parbox[c][1.2cm][c]{1.2cm}{\centering \textcolor{white}{18.8}} & \cellcolor[rgb]{0.259,0.252,0.525}\parbox[c][1.2cm][c]{1.2cm}{\centering \textcolor{white}{13.8}} & \cellcolor[rgb]{0.121,0.626,0.533}\parbox[c][1.2cm][c]{1.2cm}{\centering \textcolor{black}{25.6}} \\
\text{Emu} & \cellcolor[rgb]{0.145,0.519,0.557}\parbox[c][1.2cm][c]{1.2cm}{\centering \textcolor{white}{22.0}} & \cellcolor[rgb]{0.955,0.901,0.118}\parbox[c][1.2cm][c]{1.2cm}{\centering \textcolor{black}{38.9}} & \cellcolor[rgb]{0.247,0.283,0.536}\parbox[c][1.2cm][c]{1.2cm}{\centering \textcolor{white}{14.7}} & \cellcolor[rgb]{0.145,0.519,0.557}\parbox[c][1.2cm][c]{1.2cm}{\centering \textcolor{white}{22.1}} & \cellcolor[rgb]{0.164,0.471,0.558}\parbox[c][1.2cm][c]{1.2cm}{\centering \textcolor{white}{20.5}} \\
\text{Janus} & \cellcolor[rgb]{0.207,0.372,0.553}\parbox[c][1.2cm][c]{1.2cm}{\centering \textcolor{white}{17.3}} & \cellcolor[rgb]{0.271,0.020,0.347}\parbox[c][1.2cm][c]{1.2cm}{\centering \textcolor{white}{8.2}} & \cellcolor[rgb]{0.668,0.862,0.196}\parbox[c][1.2cm][c]{1.2cm}{\centering \textcolor{black}{35.4}} & \cellcolor[rgb]{0.212,0.360,0.552}\parbox[c][1.2cm][c]{1.2cm}{\centering \textcolor{white}{16.9}} & \cellcolor[rgb]{0.283,0.100,0.422}\parbox[c][1.2cm][c]{1.2cm}{\centering \textcolor{white}{9.9}} \\
\text{MMaDA} & \cellcolor[rgb]{0.189,0.411,0.556}\parbox[c][1.2cm][c]{1.2cm}{\centering \textcolor{white}{18.5}} & \cellcolor[rgb]{0.164,0.471,0.558}\parbox[c][1.2cm][c]{1.2cm}{\centering \textcolor{white}{20.5}} & \cellcolor[rgb]{0.203,0.380,0.554}\parbox[c][1.2cm][c]{1.2cm}{\centering \textcolor{white}{17.5}} & \cellcolor[rgb]{0.993,0.906,0.144}\parbox[c][1.2cm][c]{1.2cm}{\centering \textcolor{black}{39.4}} & \cellcolor[rgb]{0.281,0.084,0.407}\parbox[c][1.2cm][c]{1.2cm}{\centering \textcolor{white}{9.6}} \\
\text{Show-o2} & \cellcolor[rgb]{0.139,0.534,0.555}\parbox[c][1.2cm][c]{1.2cm}{\centering \textcolor{white}{22.5}} & \cellcolor[rgb]{0.237,0.305,0.542}\parbox[c][1.2cm][c]{1.2cm}{\centering \textcolor{white}{15.3}} & \cellcolor[rgb]{0.262,0.242,0.521}\parbox[c][1.2cm][c]{1.2cm}{\centering \textcolor{white}{13.5}} & \cellcolor[rgb]{0.267,0.005,0.329}\parbox[c][1.2cm][c]{1.2cm}{\centering \textcolor{white}{7.8}} & \cellcolor[rgb]{0.586,0.847,0.250}\parbox[c][1.2cm][c]{1.2cm}{\centering \textcolor{black}{34.4}} \\
\end{tabular}
}
        \caption{Precision (Col-normalized)}
        \label{tab:precision}
    \end{subtable}

    \caption{\textbf{Separability is uneven}: Janus and Show-o2 are most distinctive even at 100 training images/model.}
    \label{tab:combined_stats}
\end{table*}

\subsection{Are Off-the-Shelf MLLMs Capable Attributors?}

We investigate whether frontier off-the-shelf unified models themselves are able to reach the performance of a specialized classifier in detecting from which unified model a generated image originated. We evaluate GPT-5.4 and Gemini 3.1 Flash. For each unified model, we evaluate attribution accuracy for a given image out of 1000 random prompts fixed from MJHQ-30K~\citep{li2024playground} in addition to providing the MLLM with 0, 1, and 5 additional randomly selected examples for each candidate unified model that could have generated the image (see \Cref{app:mllm_eval} for full prompts and setup details).

\begin{figure}
    \centering
    \includegraphics[width=\columnwidth]{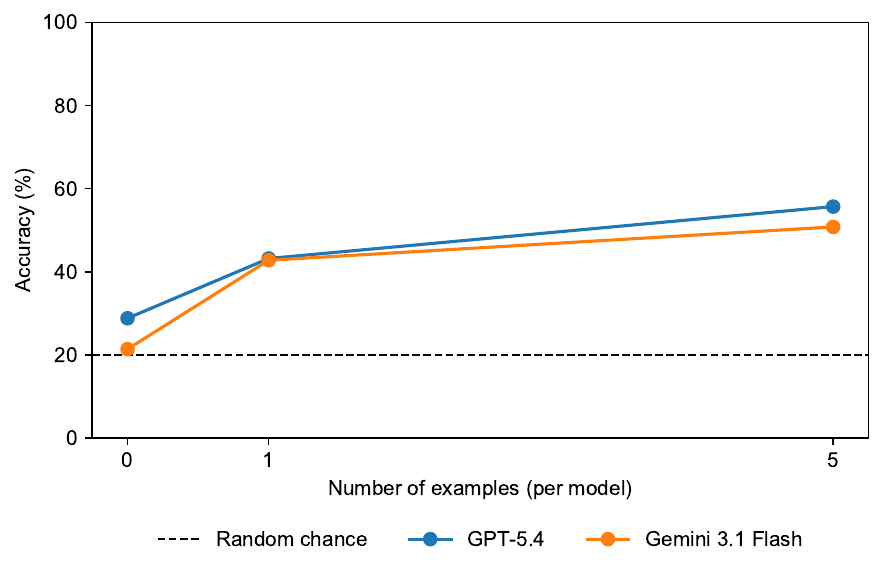}
    \caption{MLLMs plateau near 50\% accuracy even with 5-shot exemplars, well below a specialized classifier.}
    \label{fig:mllm_attribution_accuracy}
\end{figure}

As shown in \Cref{fig:mllm_attribution_accuracy}, GPT-5.4 and Gemini 3.1 Flash encounter a plateau in performance at around 50\% accuracy from 1 to 5 exemplars per candidate unified model, starting from near-chance accuracies of 28.8\% and 21.4\%, respectively. While few-shot exemplars provide a meaningful boost, frontier MLLMs plateau well below the accuracy achievable by a specialized classifier, indicating that reliable visual attribution requires deeper representational alignment than in-context prompting alone can provide.

\section{Probing Attribution Drivers: Robustness, Semantics, and Language}
\subsection{Low-Level Image Corruptions}

We use the same dataset of MJHQ-30K~\citep{li2024playground} for this experiment. This experiment evaluates the effect of low-level features of the images on the classification accuracy. We apply the corruptions to both the training and test data. We use 25K images for training and 5K images for test in all the corruption experiments. To investigate the impact of low-level features, we perform ablations using color jittering, Gaussian noise, Gaussian blur, and resizing, following standard robustness evaluation protocols~\citep{hendrycks2019benchmarking}. Despite the corruption of low-level features, accuracy remained at approximately 95\% (\Cref{tab:corruption}), suggesting that the model relies on high-level feature representations that remain robust to these perturbations.

\begin{table}[ht]
    \centering
    \begin{minipage}{0.45\textwidth}
        \centering
        \begin{tabular}{l l r}
            Corruption & Parameter & Acc. (\%) \\
            \midrule
            no corruption  & n/a            & 99.9 \\
            Color jitter   & strength 1     & 94.4 \\
            Color jitter   & strength 2     & 95.2 \\
            Gaussian noise & std 0.2        & 96.3 \\
            Gaussian noise & std 0.3        & 95.2 \\
            Gaussian blur  & radius 3       & 99.4 \\
            Gaussian blur  & radius 5       & 98.7 \\
            Resize         & $64 \times 64$ & 96.6 \\
            Resize         & $32 \times 32$ & 85.2 \\
        \end{tabular}
        \caption{Low-level corruptions barely dent accuracy, which stays near 95\% across all perturbations.}
        \label{tab:corruption}
    \end{minipage}
    \hfill
    \begin{minipage}{0.45\textwidth}
    \centering
    \begin{tabular}{l r}
        Transformation & Acc. (\%) \\
        \midrule
        no transformation     & 99.9 \\
        Depth                 & 83.2 \\
        SAM                   & 79.2 \\
        Random pixel shuffle  & 72.7 \\
    \end{tabular}
    \caption{Structural cues contribute to separability but don't fully explain it. Accuracy drops to 72--83\% under depth, segmentation, and pixel shuffle.}
    \label{tab:transformation}
\end{minipage}

\end{table}

\subsection{Structural Perturbation}

We use the same dataset of MJHQ-30K~\citep{li2024playground} for this experiment. This experiment evaluates the effect of structural bias of the images on the classification accuracy. For all the perturbations, we applied them to the training data and the test data. We used 25K images for training and 5K images for test in all the experiments. We transformed images using Depth-Anything-V2~\citep{yang2024depthv2} to make a depth map of images and Segment-Anything Model~\citep{kirillov2023segment} to apply image segmentation on images. We make a depth map to isolate the 3D structural information while removing color/texture and segmentation to isolate object-level structure and boundaries. We also apply randomized pixel shuffle to see the specificity of color distribution in each model.

With depth and segment transformation, the accuracy drops to around 80\% (\Cref{tab:transformation}). This means that the images from each model have some characteristics in depth-map and segments, but these elements alone are not enough to classify the images. Similarly, the accuracy of 72\% with random pixel shuffle suggests that the color distribution of images is different from one model to another, but that cannot explain all the differences between models.

\subsection{Out-of-Distribution Semantic Generalization}
\begin{table*}[t]
\centering
\small
\begin{tabular}{llllll}
\hline
Example & Animals & Vehicles & Arts and Works & Landscapes & Food and Drinks \\
\hline
1 & Platypus & Asphalt Paver & Oil Impasto & Geyser & Wheat \\
2 & Hare & Limousine & Trading Card Illustration & Bayou & Corn  \\
3 & Camel & Aircraft carrier & Architectural Photo Study & Levee & Strawberry  \\
4 & Tern & Horse cart & Glassblown Vessel Form & Atoll & Edamame \\
5 & Stingray & Mobile crane & Quilted Art Wall Piece & River Canyon & Feta \\
6 & Turtle & Harbor tug & Album Cover Artwork & Lush Forest & Brazil Nut  \\
7 & Firefly & Gas balloon & Surreal Collage Poster & Light Fog & Cinnamon  \\
8 & Tarantula & Bicycle & Pixel Art Tileset Map & Bog & Hummus \\
9 & Clam & Tow Truck & Spray Paint Street Mural & Butte & Turkish Coffee \\
10 & Starfish & Rowboat & Ad Campaign Storyboard & Seamount & Kombucha  \\
\hline
\end{tabular}
\end{table*}

\begin{table*}[t]
\ContinuedFloat
\centering
\small
\begin{tabular}{llllll}
\hline
Ex. & Clothing & Interior Spaces & Household Items & Buildings & People \\
\hline
1 & Jumpsuit & Basement & Freezer & Cabin & Doctor \\
2 & Fanny Pack & Studio & Air Conditioner & Skyscraper & Professor \\
3 & Jeans & Arcade Room & Printer & Convention Center & Sheriff \\
4 & Midi Skirt & X-Ray Room & Rug & Data Center Building & Musician \\
5 & Cardigan & Altar Area & Clock & Power Plant & Journalist \\
6 & Baseball Cap & Concert Hall & Mug & Water Treatment Plant & Engineer \\
7 & Hijab & Indoor Tennis Court & Fork & City Hall & DJ \\
8 & Sandals & Bowling Alley & Toolbox & Fire Station & Court Clerk \\
9 & Hazmat Suit & Baggage Claim & Broom & Hotel & Ventriloquist \\
10 & Utility Vest & Steam Room & Napkins & Supermarket & Bookkeeper \\
\hline
\end{tabular}
\vspace{0.8em}
    \captionof{table}{Ten representative examples for each domain in the 10D dataset.}
    \label{tab:10d_examples}
\end{table*}

\subsubsection{10D Dataset Creation}
We construct prompts that are mutually exclusive across domains and collectively exhaustive within each domain, so we define ten semantic domains: animals, vehicles, arts and works, landscapes, foods and drinks, clothing, interior spaces, household items, buildings, and people, chosen to represent broad, realistic visual concepts. Within each domain, we generate 300 prompts, yielding 3K prompts in total. We call this set of prompts the 10D (10 Domains) Dataset. These prompts are minimal, capturing the concept rather than subjective or stylistic phrases. We avoid adjectives, viewpoints, artistic styles, and narrative elements to eliminate subjective variance and ensure that prompts capture only the core concept being represented. Some examples are shown in \Cref{tab:10d_examples}.

Prompt generation is first done by asking an LLM to propose subcategories within each domain, and then we query the LLM for concepts within each subcategory. All concepts are reviewed to confirm that they fit their respective domain, do not overlap with other domains, and remain as objective as possible. This construction allows us to yield an OOD evaluation setting in which separability is driven by domain structure, enabling us to isolate model behavior at the conceptual level.

\subsubsection{Out-of-Distribution Experiment Trends}
This experiment evaluates whether the classifier's performance is driven by the underlying semantic content of the images or by spurious distributional trends and stylistic biases. We treat each of the 10 semantic domains in 10D as a separate OOD setting. For each domain, we train a classifier to predict the generating unified model using 200 randomly sampled images per unified model from that domain. We then evaluate each domain classifier on a held-out test set of 100 images per unified model from every domain. This yields a 10×10 accuracy matrix, where rows correspond to the domain used for training and columns to the domain used for evaluation.

The cross-domain accuracy is shown in \Cref{tab:ood_heatmap}. In order to establish a baseline, we trained 10 classifiers mixing the 10 domains with the same number of training images, resulting in an average accuracy of 46.7\%. As expected, the classifier trained on one domain and evaluated on the same domain (the diagonal of \Cref{tab:ood_heatmap}) achieves higher accuracies, except for arts and works. It achieved 45.7\% accuracy, which is lower than the baseline, and this could be explained by a wide variety of art objects, such as landscapes, portraits, and logos. Also, the matrix is not symmetric. For example, the accuracy of a classifier trained on vehicles and evaluated on food and drinks is 24.4\%, but the accuracy of a classifier trained on food and drinks and evaluated on vehicles is 39.1\%.

Although the prompts we provided are contained in a single domain and are independent from each other, it is possible that the images produced by the models are more likely to contain some domains than others, depending on the prompts. Therefore, we gave images to an image understanding model, Qwen3-VL~\citep{bai2025qwen3vltechnicalreport}, a visual language model built on the visual instruction tuning paradigm~\citep{liu2023visual}, and asked ``In the image, do you see \{question\_domain\}? Answer the question with just yes or no.'' The result is shown in \Cref{tab:domain_freq}. Surprisingly, there is not much correlation between the frequency and accuracy, which thus points toward the idea that the classification is not leveraging semantic content. Rather, it is likely relying on non-semantic, model-specific visual cues that are independent of the domain semantics.

\begin{table}[t]
\centering
\resizebox{\columnwidth}{!}{
\begin{tabular}{lw{c}{1.2cm}w{c}{1.2cm}w{c}{1.2cm}w{c}{1.2cm}w{c}{1.2cm}w{c}{1.2cm}w{c}{1.2cm}w{c}{1.2cm}w{c}{1.2cm}w{c}{1.2cm}}
\diagbox[]{\textbf{Train}}{\textbf{Eval}} & \rotatebox{45}{animals} & \rotatebox{45}{arts and works} & \rotatebox{45}{buildings} & \rotatebox{45}{clothing} & \rotatebox{45}{food and drinks} & \rotatebox{45}{household items} & \rotatebox{45}{interior spaces} & \rotatebox{45}{landscapes} & \rotatebox{45}{people} & \rotatebox{45}{vehicles} \\
\text{animals} & \cellcolor[rgb]{0.289,0.758,0.428}\parbox[c][1.2cm][c]{1.2cm}{\centering \textcolor{black}{58.6}} & \cellcolor[rgb]{0.218,0.347,0.550}\parbox[c][1.2cm][c]{1.2cm}{\centering \textcolor{white}{37.5}} & \cellcolor[rgb]{0.198,0.392,0.555}\parbox[c][1.2cm][c]{1.2cm}{\centering \textcolor{white}{39.6}} & \cellcolor[rgb]{0.203,0.380,0.554}\parbox[c][1.2cm][c]{1.2cm}{\centering \textcolor{white}{39.0}} & \cellcolor[rgb]{0.165,0.467,0.558}\parbox[c][1.2cm][c]{1.2cm}{\centering \textcolor{white}{43.3}} & \cellcolor[rgb]{0.174,0.445,0.558}\parbox[c][1.2cm][c]{1.2cm}{\centering \textcolor{white}{42.3}} & \cellcolor[rgb]{0.152,0.501,0.558}\parbox[c][1.2cm][c]{1.2cm}{\centering \textcolor{white}{45.0}} & \cellcolor[rgb]{0.230,0.322,0.546}\parbox[c][1.2cm][c]{1.2cm}{\centering \textcolor{white}{36.4}} & \cellcolor[rgb]{0.179,0.434,0.557}\parbox[c][1.2cm][c]{1.2cm}{\centering \textcolor{white}{41.7}} & \cellcolor[rgb]{0.243,0.292,0.539}\parbox[c][1.2cm][c]{1.2cm}{\centering \textcolor{white}{35.1}} \\
\text{arts and works} & \cellcolor[rgb]{0.228,0.327,0.547}\parbox[c][1.2cm][c]{1.2cm}{\centering \textcolor{white}{36.6}} & \cellcolor[rgb]{0.146,0.515,0.557}\parbox[c][1.2cm][c]{1.2cm}{\centering \textcolor{white}{45.7}} & \cellcolor[rgb]{0.209,0.368,0.553}\parbox[c][1.2cm][c]{1.2cm}{\centering \textcolor{white}{38.6}} & \cellcolor[rgb]{0.184,0.422,0.557}\parbox[c][1.2cm][c]{1.2cm}{\centering \textcolor{white}{41.1}} & \cellcolor[rgb]{0.269,0.219,0.510}\parbox[c][1.2cm][c]{1.2cm}{\centering \textcolor{white}{32.1}} & \cellcolor[rgb]{0.245,0.288,0.537}\parbox[c][1.2cm][c]{1.2cm}{\centering \textcolor{white}{34.9}} & \cellcolor[rgb]{0.170,0.456,0.558}\parbox[c][1.2cm][c]{1.2cm}{\centering \textcolor{white}{42.7}} & \cellcolor[rgb]{0.283,0.136,0.453}\parbox[c][1.2cm][c]{1.2cm}{\centering \textcolor{white}{29.0}} & \cellcolor[rgb]{0.150,0.504,0.557}\parbox[c][1.2cm][c]{1.2cm}{\centering \textcolor{white}{45.3}} & \cellcolor[rgb]{0.267,0.228,0.514}\parbox[c][1.2cm][c]{1.2cm}{\centering \textcolor{white}{32.6}} \\
\text{buildings} & \cellcolor[rgb]{0.203,0.380,0.554}\parbox[c][1.2cm][c]{1.2cm}{\centering \textcolor{white}{39.1}} & \cellcolor[rgb]{0.228,0.327,0.547}\parbox[c][1.2cm][c]{1.2cm}{\centering \textcolor{white}{36.6}} & \cellcolor[rgb]{0.936,0.899,0.108}\parbox[c][1.2cm][c]{1.2cm}{\centering \textcolor{black}{71.1}} & \cellcolor[rgb]{0.222,0.339,0.549}\parbox[c][1.2cm][c]{1.2cm}{\centering \textcolor{white}{37.1}} & \cellcolor[rgb]{0.280,0.171,0.480}\parbox[c][1.2cm][c]{1.2cm}{\centering \textcolor{white}{30.2}} & \cellcolor[rgb]{0.220,0.343,0.549}\parbox[c][1.2cm][c]{1.2cm}{\centering \textcolor{white}{37.4}} & \cellcolor[rgb]{0.132,0.655,0.520}\parbox[c][1.2cm][c]{1.2cm}{\centering \textcolor{black}{53.0}} & \cellcolor[rgb]{0.155,0.493,0.558}\parbox[c][1.2cm][c]{1.2cm}{\centering \textcolor{white}{44.7}} & \cellcolor[rgb]{0.125,0.640,0.527}\parbox[c][1.2cm][c]{1.2cm}{\centering \textcolor{black}{52.1}} & \cellcolor[rgb]{0.122,0.589,0.546}\parbox[c][1.2cm][c]{1.2cm}{\centering \textcolor{black}{49.6}} \\
\text{clothing} & \cellcolor[rgb]{0.181,0.430,0.557}\parbox[c][1.2cm][c]{1.2cm}{\centering \textcolor{white}{41.6}} & \cellcolor[rgb]{0.236,0.310,0.543}\parbox[c][1.2cm][c]{1.2cm}{\centering \textcolor{white}{35.9}} & \cellcolor[rgb]{0.275,0.195,0.496}\parbox[c][1.2cm][c]{1.2cm}{\centering \textcolor{white}{31.3}} & \cellcolor[rgb]{0.404,0.800,0.363}\parbox[c][1.2cm][c]{1.2cm}{\centering \textcolor{black}{61.1}} & \cellcolor[rgb]{0.177,0.438,0.558}\parbox[c][1.2cm][c]{1.2cm}{\centering \textcolor{white}{41.9}} & \cellcolor[rgb]{0.150,0.504,0.557}\parbox[c][1.2cm][c]{1.2cm}{\centering \textcolor{white}{45.1}} & \cellcolor[rgb]{0.220,0.343,0.549}\parbox[c][1.2cm][c]{1.2cm}{\centering \textcolor{white}{37.3}} & \cellcolor[rgb]{0.254,0.265,0.530}\parbox[c][1.2cm][c]{1.2cm}{\centering \textcolor{white}{34.0}} & \cellcolor[rgb]{0.186,0.419,0.557}\parbox[c][1.2cm][c]{1.2cm}{\centering \textcolor{white}{41.0}} & \cellcolor[rgb]{0.275,0.195,0.496}\parbox[c][1.2cm][c]{1.2cm}{\centering \textcolor{white}{31.3}} \\
\text{food and drinks} & \cellcolor[rgb]{0.121,0.596,0.544}\parbox[c][1.2cm][c]{1.2cm}{\centering \textcolor{black}{50.0}} & \cellcolor[rgb]{0.234,0.314,0.544}\parbox[c][1.2cm][c]{1.2cm}{\centering \textcolor{white}{36.0}} & \cellcolor[rgb]{0.184,0.422,0.557}\parbox[c][1.2cm][c]{1.2cm}{\centering \textcolor{white}{41.1}} & \cellcolor[rgb]{0.126,0.571,0.550}\parbox[c][1.2cm][c]{1.2cm}{\centering \textcolor{black}{48.6}} & \cellcolor[rgb]{0.606,0.851,0.237}\parbox[c][1.2cm][c]{1.2cm}{\centering \textcolor{black}{65.1}} & \cellcolor[rgb]{0.167,0.464,0.558}\parbox[c][1.2cm][c]{1.2cm}{\centering \textcolor{white}{43.1}} & \cellcolor[rgb]{0.184,0.422,0.557}\parbox[c][1.2cm][c]{1.2cm}{\centering \textcolor{white}{41.1}} & \cellcolor[rgb]{0.176,0.441,0.558}\parbox[c][1.2cm][c]{1.2cm}{\centering \textcolor{white}{42.1}} & \cellcolor[rgb]{0.126,0.571,0.550}\parbox[c][1.2cm][c]{1.2cm}{\centering \textcolor{black}{48.6}} & \cellcolor[rgb]{0.203,0.380,0.554}\parbox[c][1.2cm][c]{1.2cm}{\centering \textcolor{white}{39.1}} \\
\text{household items} & \cellcolor[rgb]{0.149,0.508,0.557}\parbox[c][1.2cm][c]{1.2cm}{\centering \textcolor{white}{45.4}} & \cellcolor[rgb]{0.207,0.372,0.553}\parbox[c][1.2cm][c]{1.2cm}{\centering \textcolor{white}{38.6}} & \cellcolor[rgb]{0.194,0.399,0.556}\parbox[c][1.2cm][c]{1.2cm}{\centering \textcolor{white}{40.0}} & \cellcolor[rgb]{0.128,0.648,0.523}\parbox[c][1.2cm][c]{1.2cm}{\centering \textcolor{black}{52.6}} & \cellcolor[rgb]{0.139,0.534,0.555}\parbox[c][1.2cm][c]{1.2cm}{\centering \textcolor{white}{46.7}} & \cellcolor[rgb]{0.191,0.708,0.482}\parbox[c][1.2cm][c]{1.2cm}{\centering \textcolor{black}{55.7}} & \cellcolor[rgb]{0.129,0.563,0.551}\parbox[c][1.2cm][c]{1.2cm}{\centering \textcolor{white}{48.3}} & \cellcolor[rgb]{0.275,0.195,0.496}\parbox[c][1.2cm][c]{1.2cm}{\centering \textcolor{white}{31.3}} & \cellcolor[rgb]{0.130,0.560,0.552}\parbox[c][1.2cm][c]{1.2cm}{\centering \textcolor{white}{48.0}} & \cellcolor[rgb]{0.209,0.368,0.553}\parbox[c][1.2cm][c]{1.2cm}{\centering \textcolor{white}{38.4}} \\
\text{interior spaces} & \cellcolor[rgb]{0.189,0.411,0.556}\parbox[c][1.2cm][c]{1.2cm}{\centering \textcolor{white}{40.6}} & \cellcolor[rgb]{0.170,0.456,0.558}\parbox[c][1.2cm][c]{1.2cm}{\centering \textcolor{white}{42.7}} & \cellcolor[rgb]{0.140,0.666,0.513}\parbox[c][1.2cm][c]{1.2cm}{\centering \textcolor{black}{53.4}} & \cellcolor[rgb]{0.232,0.318,0.545}\parbox[c][1.2cm][c]{1.2cm}{\centering \textcolor{white}{36.3}} & \cellcolor[rgb]{0.198,0.392,0.555}\parbox[c][1.2cm][c]{1.2cm}{\centering \textcolor{white}{39.6}} & \cellcolor[rgb]{0.156,0.490,0.558}\parbox[c][1.2cm][c]{1.2cm}{\centering \textcolor{white}{44.4}} & \cellcolor[rgb]{0.678,0.864,0.190}\parbox[c][1.2cm][c]{1.2cm}{\centering \textcolor{black}{66.4}} & \cellcolor[rgb]{0.189,0.411,0.556}\parbox[c][1.2cm][c]{1.2cm}{\centering \textcolor{white}{40.6}} & \cellcolor[rgb]{0.123,0.637,0.529}\parbox[c][1.2cm][c]{1.2cm}{\centering \textcolor{black}{52.0}} & \cellcolor[rgb]{0.134,0.549,0.554}\parbox[c][1.2cm][c]{1.2cm}{\centering \textcolor{white}{47.6}} \\
\text{landscapes} & \cellcolor[rgb]{0.259,0.252,0.525}\parbox[c][1.2cm][c]{1.2cm}{\centering \textcolor{white}{33.4}} & \cellcolor[rgb]{0.241,0.296,0.540}\parbox[c][1.2cm][c]{1.2cm}{\centering \textcolor{white}{35.3}} & \cellcolor[rgb]{0.152,0.501,0.558}\parbox[c][1.2cm][c]{1.2cm}{\centering \textcolor{white}{45.0}} & \cellcolor[rgb]{0.250,0.274,0.533}\parbox[c][1.2cm][c]{1.2cm}{\centering \textcolor{white}{34.3}} & \cellcolor[rgb]{0.243,0.292,0.539}\parbox[c][1.2cm][c]{1.2cm}{\centering \textcolor{white}{35.2}} & \cellcolor[rgb]{0.257,0.256,0.527}\parbox[c][1.2cm][c]{1.2cm}{\centering \textcolor{white}{33.6}} & \cellcolor[rgb]{0.181,0.430,0.557}\parbox[c][1.2cm][c]{1.2cm}{\centering \textcolor{white}{41.4}} & \cellcolor[rgb]{0.191,0.708,0.482}\parbox[c][1.2cm][c]{1.2cm}{\centering \textcolor{black}{55.7}} & \cellcolor[rgb]{0.165,0.467,0.558}\parbox[c][1.2cm][c]{1.2cm}{\centering \textcolor{white}{43.3}} & \cellcolor[rgb]{0.181,0.430,0.557}\parbox[c][1.2cm][c]{1.2cm}{\centering \textcolor{white}{41.4}} \\
\text{people} & \cellcolor[rgb]{0.222,0.339,0.549}\parbox[c][1.2cm][c]{1.2cm}{\centering \textcolor{white}{37.1}} & \cellcolor[rgb]{0.220,0.343,0.549}\parbox[c][1.2cm][c]{1.2cm}{\centering \textcolor{white}{37.3}} & \cellcolor[rgb]{0.152,0.501,0.558}\parbox[c][1.2cm][c]{1.2cm}{\centering \textcolor{white}{45.0}} & \cellcolor[rgb]{0.226,0.331,0.547}\parbox[c][1.2cm][c]{1.2cm}{\centering \textcolor{white}{36.9}} & \cellcolor[rgb]{0.218,0.347,0.550}\parbox[c][1.2cm][c]{1.2cm}{\centering \textcolor{white}{37.5}} & \cellcolor[rgb]{0.216,0.352,0.551}\parbox[c][1.2cm][c]{1.2cm}{\centering \textcolor{white}{37.7}} & \cellcolor[rgb]{0.122,0.633,0.530}\parbox[c][1.2cm][c]{1.2cm}{\centering \textcolor{black}{51.7}} & \cellcolor[rgb]{0.234,0.314,0.544}\parbox[c][1.2cm][c]{1.2cm}{\centering \textcolor{white}{36.0}} & \cellcolor[rgb]{0.993,0.906,0.144}\parbox[c][1.2cm][c]{1.2cm}{\centering \textcolor{black}{72.3}} & \cellcolor[rgb]{0.173,0.449,0.558}\parbox[c][1.2cm][c]{1.2cm}{\centering \textcolor{white}{42.4}} \\
\text{vehicles} & \cellcolor[rgb]{0.282,0.151,0.465}\parbox[c][1.2cm][c]{1.2cm}{\centering \textcolor{white}{29.4}} & \cellcolor[rgb]{0.234,0.314,0.544}\parbox[c][1.2cm][c]{1.2cm}{\centering \textcolor{white}{36.0}} & \cellcolor[rgb]{0.125,0.640,0.527}\parbox[c][1.2cm][c]{1.2cm}{\centering \textcolor{black}{52.1}} & \cellcolor[rgb]{0.257,0.256,0.527}\parbox[c][1.2cm][c]{1.2cm}{\centering \textcolor{white}{33.7}} & \cellcolor[rgb]{0.267,0.005,0.329}\parbox[c][1.2cm][c]{1.2cm}{\centering \textcolor{white}{24.4}} & \cellcolor[rgb]{0.283,0.105,0.427}\parbox[c][1.2cm][c]{1.2cm}{\centering \textcolor{white}{27.9}} & \cellcolor[rgb]{0.176,0.441,0.558}\parbox[c][1.2cm][c]{1.2cm}{\centering \textcolor{white}{42.0}} & \cellcolor[rgb]{0.186,0.419,0.557}\parbox[c][1.2cm][c]{1.2cm}{\centering \textcolor{white}{41.0}} & \cellcolor[rgb]{0.138,0.537,0.555}\parbox[c][1.2cm][c]{1.2cm}{\centering \textcolor{white}{46.9}} & \cellcolor[rgb]{0.555,0.840,0.269}\parbox[c][1.2cm][c]{1.2cm}{\centering \textcolor{black}{64.1}} \\
\end{tabular}
}
\caption{\textbf{Cross-domain accuracy stays well above chance, showing model-specific signals generalize across semantic domains.} Cell (X, Y) is the accuracy of a classifier trained on domain X and evaluated on domain Y across 7 models; random chance is $100/7 \approx 14.3\%$.}
\label{tab:ood_heatmap}

\end{table}

\begin{table}[t]
\centering
\resizebox{\columnwidth}{!}{
\begin{tabular}{lw{c}{1.2cm}w{c}{1.2cm}w{c}{1.2cm}w{c}{1.2cm}w{c}{1.2cm}w{c}{1.2cm}w{c}{1.2cm}w{c}{1.2cm}w{c}{1.2cm}w{c}{1.2cm}}
\diagbox{\textbf{Orig.}}{\textbf{Quest.}} & \rotatebox{45}{animals} & \rotatebox{45}{arts and works} & \rotatebox{45}{buildings} & \rotatebox{45}{clothing} & \rotatebox{45}{food and drinks} & \rotatebox{45}{household items} & \rotatebox{45}{interior spaces} & \rotatebox{45}{landscapes} & \rotatebox{45}{people} & \rotatebox{45}{vehicles} \\
\text{animals} & \cellcolor[rgb]{0.946,0.900,0.113}\parbox[c][1.2cm][c]{1.2cm}{\centering \textcolor{black}{96.7}} & \cellcolor[rgb]{0.281,0.084,0.407}\parbox[c][1.2cm][c]{1.2cm}{\centering \textcolor{white}{5.7}} & \cellcolor[rgb]{0.279,0.062,0.387}\parbox[c][1.2cm][c]{1.2cm}{\centering \textcolor{white}{4.0}} & \cellcolor[rgb]{0.276,0.044,0.370}\parbox[c][1.2cm][c]{1.2cm}{\centering \textcolor{white}{3.0}} & \cellcolor[rgb]{0.282,0.095,0.417}\parbox[c][1.2cm][c]{1.2cm}{\centering \textcolor{white}{6.3}} & \cellcolor[rgb]{0.276,0.044,0.370}\parbox[c][1.2cm][c]{1.2cm}{\centering \textcolor{white}{3.0}} & \cellcolor[rgb]{0.283,0.100,0.422}\parbox[c][1.2cm][c]{1.2cm}{\centering \textcolor{white}{6.7}} & \cellcolor[rgb]{0.171,0.694,0.494}\parbox[c][1.2cm][c]{1.2cm}{\centering \textcolor{black}{63.3}} & \cellcolor[rgb]{0.274,0.031,0.359}\parbox[c][1.2cm][c]{1.2cm}{\centering \textcolor{white}{2.0}} & \cellcolor[rgb]{0.269,0.010,0.335}\parbox[c][1.2cm][c]{1.2cm}{\centering \textcolor{white}{0.7}} \\
\text{arts and works} & \cellcolor[rgb]{0.261,0.247,0.523}\parbox[c][1.2cm][c]{1.2cm}{\centering \textcolor{white}{18.3}} & \cellcolor[rgb]{0.565,0.842,0.263}\parbox[c][1.2cm][c]{1.2cm}{\centering \textcolor{black}{82.4}} & \cellcolor[rgb]{0.236,0.310,0.543}\parbox[c][1.2cm][c]{1.2cm}{\centering \textcolor{white}{23.9}} & \cellcolor[rgb]{0.230,0.322,0.546}\parbox[c][1.2cm][c]{1.2cm}{\centering \textcolor{white}{24.9}} & \cellcolor[rgb]{0.281,0.079,0.402}\parbox[c][1.2cm][c]{1.2cm}{\centering \textcolor{white}{5.3}} & \cellcolor[rgb]{0.232,0.318,0.545}\parbox[c][1.2cm][c]{1.2cm}{\centering \textcolor{white}{24.6}} & \cellcolor[rgb]{0.191,0.407,0.556}\parbox[c][1.2cm][c]{1.2cm}{\centering \textcolor{white}{32.9}} & \cellcolor[rgb]{0.205,0.376,0.554}\parbox[c][1.2cm][c]{1.2cm}{\centering \textcolor{white}{29.9}} & \cellcolor[rgb]{0.228,0.327,0.547}\parbox[c][1.2cm][c]{1.2cm}{\centering \textcolor{white}{25.2}} & \cellcolor[rgb]{0.283,0.111,0.432}\parbox[c][1.2cm][c]{1.2cm}{\centering \textcolor{white}{7.6}} \\
\text{buildings} & \cellcolor[rgb]{0.267,0.005,0.329}\parbox[c][1.2cm][c]{1.2cm}{\centering \textcolor{white}{0.3}} & \cellcolor[rgb]{0.281,0.156,0.469}\parbox[c][1.2cm][c]{1.2cm}{\centering \textcolor{white}{11.0}} & \cellcolor[rgb]{0.906,0.895,0.098}\parbox[c][1.2cm][c]{1.2cm}{\centering \textcolor{black}{95.3}} & \cellcolor[rgb]{0.148,0.512,0.557}\parbox[c][1.2cm][c]{1.2cm}{\centering \textcolor{white}{43.7}} & \cellcolor[rgb]{0.275,0.038,0.365}\parbox[c][1.2cm][c]{1.2cm}{\centering \textcolor{white}{2.7}} & \cellcolor[rgb]{0.283,0.100,0.422}\parbox[c][1.2cm][c]{1.2cm}{\centering \textcolor{white}{6.7}} & \cellcolor[rgb]{0.230,0.322,0.546}\parbox[c][1.2cm][c]{1.2cm}{\centering \textcolor{white}{25.0}} & \cellcolor[rgb]{0.132,0.655,0.520}\parbox[c][1.2cm][c]{1.2cm}{\centering \textcolor{black}{59.0}} & \cellcolor[rgb]{0.123,0.585,0.547}\parbox[c][1.2cm][c]{1.2cm}{\centering \textcolor{black}{51.7}} & \cellcolor[rgb]{0.158,0.486,0.558}\parbox[c][1.2cm][c]{1.2cm}{\centering \textcolor{white}{41.3}} \\
\text{clothing} & \cellcolor[rgb]{0.279,0.062,0.387}\parbox[c][1.2cm][c]{1.2cm}{\centering \textcolor{white}{4.0}} & \cellcolor[rgb]{0.282,0.146,0.462}\parbox[c][1.2cm][c]{1.2cm}{\centering \textcolor{white}{10.3}} & \cellcolor[rgb]{0.220,0.343,0.549}\parbox[c][1.2cm][c]{1.2cm}{\centering \textcolor{white}{27.0}} & \cellcolor[rgb]{0.431,0.808,0.346}\parbox[c][1.2cm][c]{1.2cm}{\centering \textcolor{black}{77.3}} & \cellcolor[rgb]{0.283,0.105,0.427}\parbox[c][1.2cm][c]{1.2cm}{\centering \textcolor{white}{7.3}} & \cellcolor[rgb]{0.211,0.364,0.552}\parbox[c][1.2cm][c]{1.2cm}{\centering \textcolor{white}{28.7}} & \cellcolor[rgb]{0.150,0.504,0.557}\parbox[c][1.2cm][c]{1.2cm}{\centering \textcolor{white}{43.3}} & \cellcolor[rgb]{0.187,0.415,0.557}\parbox[c][1.2cm][c]{1.2cm}{\centering \textcolor{white}{33.7}} & \cellcolor[rgb]{0.134,0.549,0.554}\parbox[c][1.2cm][c]{1.2cm}{\centering \textcolor{white}{47.7}} & \cellcolor[rgb]{0.281,0.079,0.402}\parbox[c][1.2cm][c]{1.2cm}{\centering \textcolor{white}{5.3}} \\
\text{food and drinks} & \cellcolor[rgb]{0.276,0.044,0.370}\parbox[c][1.2cm][c]{1.2cm}{\centering \textcolor{white}{3.0}} & \cellcolor[rgb]{0.267,0.005,0.329}\parbox[c][1.2cm][c]{1.2cm}{\centering \textcolor{white}{0.3}} & \cellcolor[rgb]{0.280,0.068,0.392}\parbox[c][1.2cm][c]{1.2cm}{\centering \textcolor{white}{4.3}} & \cellcolor[rgb]{0.276,0.044,0.370}\parbox[c][1.2cm][c]{1.2cm}{\centering \textcolor{white}{3.0}} & \cellcolor[rgb]{0.762,0.876,0.137}\parbox[c][1.2cm][c]{1.2cm}{\centering \textcolor{black}{90.0}} & \cellcolor[rgb]{0.126,0.644,0.525}\parbox[c][1.2cm][c]{1.2cm}{\centering \textcolor{black}{57.8}} & \cellcolor[rgb]{0.129,0.563,0.551}\parbox[c][1.2cm][c]{1.2cm}{\centering \textcolor{white}{49.2}} & \cellcolor[rgb]{0.268,0.224,0.512}\parbox[c][1.2cm][c]{1.2cm}{\centering \textcolor{white}{16.3}} & \cellcolor[rgb]{0.280,0.068,0.392}\parbox[c][1.2cm][c]{1.2cm}{\centering \textcolor{white}{4.3}} & \cellcolor[rgb]{0.269,0.010,0.335}\parbox[c][1.2cm][c]{1.2cm}{\centering \textcolor{white}{0.7}} \\
\text{household items} & \cellcolor[rgb]{0.276,0.044,0.370}\parbox[c][1.2cm][c]{1.2cm}{\centering \textcolor{white}{3.0}} & \cellcolor[rgb]{0.271,0.214,0.507}\parbox[c][1.2cm][c]{1.2cm}{\centering \textcolor{white}{15.7}} & \cellcolor[rgb]{0.280,0.171,0.480}\parbox[c][1.2cm][c]{1.2cm}{\centering \textcolor{white}{12.3}} & \cellcolor[rgb]{0.279,0.175,0.483}\parbox[c][1.2cm][c]{1.2cm}{\centering \textcolor{white}{12.7}} & \cellcolor[rgb]{0.222,0.339,0.549}\parbox[c][1.2cm][c]{1.2cm}{\centering \textcolor{white}{26.3}} & \cellcolor[rgb]{0.741,0.873,0.150}\parbox[c][1.2cm][c]{1.2cm}{\centering \textcolor{black}{89.3}} & \cellcolor[rgb]{0.596,0.849,0.243}\parbox[c][1.2cm][c]{1.2cm}{\centering \textcolor{black}{83.7}} & \cellcolor[rgb]{0.271,0.214,0.507}\parbox[c][1.2cm][c]{1.2cm}{\centering \textcolor{white}{15.7}} & \cellcolor[rgb]{0.282,0.090,0.412}\parbox[c][1.2cm][c]{1.2cm}{\centering \textcolor{white}{6.0}} & \cellcolor[rgb]{0.270,0.015,0.341}\parbox[c][1.2cm][c]{1.2cm}{\centering \textcolor{white}{1.0}} \\
\text{interior spaces} & \cellcolor[rgb]{0.271,0.020,0.347}\parbox[c][1.2cm][c]{1.2cm}{\centering \textcolor{white}{1.3}} & \cellcolor[rgb]{0.211,0.364,0.552}\parbox[c][1.2cm][c]{1.2cm}{\centering \textcolor{white}{29.0}} & \cellcolor[rgb]{0.142,0.526,0.556}\parbox[c][1.2cm][c]{1.2cm}{\centering \textcolor{white}{45.3}} & \cellcolor[rgb]{0.214,0.722,0.470}\parbox[c][1.2cm][c]{1.2cm}{\centering \textcolor{black}{66.3}} & \cellcolor[rgb]{0.283,0.126,0.445}\parbox[c][1.2cm][c]{1.2cm}{\centering \textcolor{white}{8.7}} & \cellcolor[rgb]{0.257,0.256,0.527}\parbox[c][1.2cm][c]{1.2cm}{\centering \textcolor{white}{19.3}} & \cellcolor[rgb]{0.876,0.891,0.095}\parbox[c][1.2cm][c]{1.2cm}{\centering \textcolor{black}{94.3}} & \cellcolor[rgb]{0.278,0.180,0.487}\parbox[c][1.2cm][c]{1.2cm}{\centering \textcolor{white}{13.0}} & \cellcolor[rgb]{0.171,0.694,0.494}\parbox[c][1.2cm][c]{1.2cm}{\centering \textcolor{black}{63.3}} & \cellcolor[rgb]{0.283,0.141,0.458}\parbox[c][1.2cm][c]{1.2cm}{\centering \textcolor{white}{10.0}} \\
\text{landscapes} & \cellcolor[rgb]{0.281,0.156,0.469}\parbox[c][1.2cm][c]{1.2cm}{\centering \textcolor{white}{11.0}} & \cellcolor[rgb]{0.280,0.166,0.476}\parbox[c][1.2cm][c]{1.2cm}{\centering \textcolor{white}{11.7}} & \cellcolor[rgb]{0.279,0.175,0.483}\parbox[c][1.2cm][c]{1.2cm}{\centering \textcolor{white}{12.7}} & \cellcolor[rgb]{0.279,0.062,0.387}\parbox[c][1.2cm][c]{1.2cm}{\centering \textcolor{white}{4.0}} & \cellcolor[rgb]{0.267,0.005,0.329}\parbox[c][1.2cm][c]{1.2cm}{\centering \textcolor{white}{0.0}} & \cellcolor[rgb]{0.267,0.005,0.329}\parbox[c][1.2cm][c]{1.2cm}{\centering \textcolor{white}{0.0}} & \cellcolor[rgb]{0.276,0.044,0.370}\parbox[c][1.2cm][c]{1.2cm}{\centering \textcolor{white}{3.0}} & \cellcolor[rgb]{0.974,0.904,0.130}\parbox[c][1.2cm][c]{1.2cm}{\centering \textcolor{black}{98.0}} & \cellcolor[rgb]{0.281,0.079,0.402}\parbox[c][1.2cm][c]{1.2cm}{\centering \textcolor{white}{5.3}} & \cellcolor[rgb]{0.276,0.044,0.370}\parbox[c][1.2cm][c]{1.2cm}{\centering \textcolor{white}{3.0}} \\
\text{people} & \cellcolor[rgb]{0.280,0.068,0.392}\parbox[c][1.2cm][c]{1.2cm}{\centering \textcolor{white}{4.3}} & \cellcolor[rgb]{0.220,0.343,0.549}\parbox[c][1.2cm][c]{1.2cm}{\centering \textcolor{white}{26.7}} & \cellcolor[rgb]{0.146,0.515,0.557}\parbox[c][1.2cm][c]{1.2cm}{\centering \textcolor{white}{44.3}} & \cellcolor[rgb]{0.993,0.906,0.144}\parbox[c][1.2cm][c]{1.2cm}{\centering \textcolor{black}{99.0}} & \cellcolor[rgb]{0.252,0.270,0.532}\parbox[c][1.2cm][c]{1.2cm}{\centering \textcolor{white}{20.3}} & \cellcolor[rgb]{0.252,0.270,0.532}\parbox[c][1.2cm][c]{1.2cm}{\centering \textcolor{white}{20.3}} & \cellcolor[rgb]{0.497,0.826,0.306}\parbox[c][1.2cm][c]{1.2cm}{\centering \textcolor{black}{79.7}} & \cellcolor[rgb]{0.279,0.175,0.483}\parbox[c][1.2cm][c]{1.2cm}{\centering \textcolor{white}{12.7}} & \cellcolor[rgb]{0.993,0.906,0.144}\parbox[c][1.2cm][c]{1.2cm}{\centering \textcolor{black}{99.0}} & \cellcolor[rgb]{0.277,0.185,0.490}\parbox[c][1.2cm][c]{1.2cm}{\centering \textcolor{white}{13.3}} \\
\text{vehicles} & \cellcolor[rgb]{0.280,0.073,0.397}\parbox[c][1.2cm][c]{1.2cm}{\centering \textcolor{white}{5.0}} & \cellcolor[rgb]{0.281,0.079,0.402}\parbox[c][1.2cm][c]{1.2cm}{\centering \textcolor{white}{5.3}} & \cellcolor[rgb]{0.179,0.434,0.557}\parbox[c][1.2cm][c]{1.2cm}{\centering \textcolor{white}{35.7}} & \cellcolor[rgb]{0.141,0.530,0.556}\parbox[c][1.2cm][c]{1.2cm}{\centering \textcolor{white}{46.0}} & \cellcolor[rgb]{0.274,0.031,0.359}\parbox[c][1.2cm][c]{1.2cm}{\centering \textcolor{white}{2.3}} & \cellcolor[rgb]{0.271,0.020,0.347}\parbox[c][1.2cm][c]{1.2cm}{\centering \textcolor{white}{1.3}} & \cellcolor[rgb]{0.274,0.031,0.359}\parbox[c][1.2cm][c]{1.2cm}{\centering \textcolor{white}{2.3}} & \cellcolor[rgb]{0.336,0.777,0.402}\parbox[c][1.2cm][c]{1.2cm}{\centering \textcolor{black}{73.0}} & \cellcolor[rgb]{0.126,0.571,0.550}\parbox[c][1.2cm][c]{1.2cm}{\centering \textcolor{black}{50.0}} & \cellcolor[rgb]{0.312,0.768,0.416}\parbox[c][1.2cm][c]{1.2cm}{\centering \textcolor{black}{71.7}} \\
\end{tabular}
}
\caption{\textbf{Domain co-occurrence shows little correlation with classification accuracy, ruling out semantic content as the primary signal.} Orig.\ is the prompt domain; Quest.\ is the domain queried in the image. Cell values are the percentage of Orig.\ images containing elements of Quest.\ (e.g., (people, animals) = 4.3\%).}
\label{tab:domain_freq}
\end{table}

\subsection{Cross-Linguistic Visual Signatures}
We use the same MJHQ-30K dataset~\citep{li2024playground} to evaluate the linguistic separability of each unified model, examining whether visual representations are language dependent. We begin by translating 1000 randomly sampled prompts using Google Translate into 4 additional languages: Spanish, Turkish, Japanese, and Simplified Chinese, chosen to cover diverse linguistic families. Using these 5 languages (including English), we train a classifier to identify the prompt language from the generated images for each unified model, and then evaluate overall accuracy on a held-out test set. We used 700 images for each language for training and 300 images for evaluation.

\begin{figure*}[ht]
    \centering
    \begin{minipage}{0.16\textwidth}\centering\small English\end{minipage}\hfill
    \begin{minipage}{0.16\textwidth}\centering\small Spanish\end{minipage}\hfill
    \begin{minipage}{0.16\textwidth}\centering\small Japanese\end{minipage}\hfill
    \begin{minipage}{0.16\textwidth}\centering\small Turkish\end{minipage}\hfill
    \begin{minipage}{0.16\textwidth}\centering\small Chinese\end{minipage}
    \vspace{2pt}
    \begin{subfigure}[b]{\textwidth}
        \centering
        \includegraphics[width=0.16\textwidth]{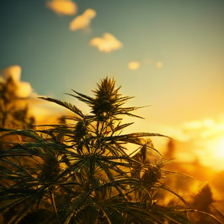}\hfill
        \includegraphics[width=0.16\textwidth]{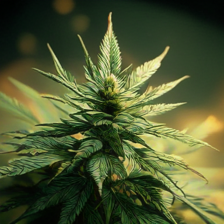}\hfill
        \includegraphics[width=0.16\textwidth]{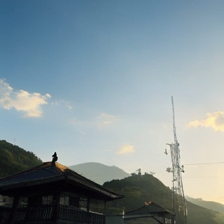}\hfill
        \includegraphics[width=0.16\textwidth]{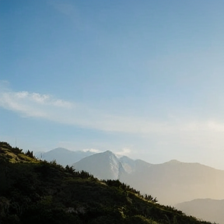}\hfill
        \includegraphics[width=0.16\textwidth]{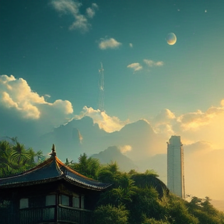}
        \caption{Janus -- ``Golden Hour, cannabis, hyper realistic, futuristic optics, highly detailed''}
        \label{fig:lang-comparison-janus}
    \end{subfigure}
    \vspace{4pt}
    \begin{subfigure}[b]{\textwidth}
        \centering
        \includegraphics[width=0.16\textwidth]{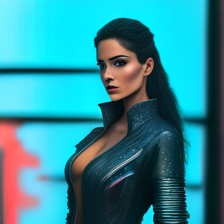}\hfill
        \includegraphics[width=0.16\textwidth]{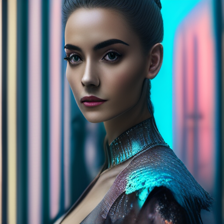}\hfill
        \includegraphics[width=0.16\textwidth]{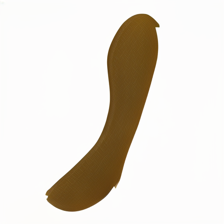}\hfill
        \includegraphics[width=0.16\textwidth]{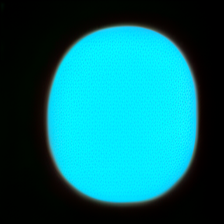}\hfill
        \includegraphics[width=0.16\textwidth]{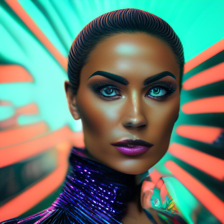}
        \caption{MMaDA -- ``the most beautiful woman in the world beautiful colombian woman wearing cyberpunk clothes, standing in the rain on a cyberpunk city street, high angle, neon lights, Use a Nikon D850 DSLR camera with a 200mm lens at F 1.2 aperture setting to isolate the subject, full body shot''}
        \label{fig:lang-comparison-mmada}
    \end{subfigure}
    \vspace{4pt}
    \begin{subfigure}[b]{\textwidth}
        \centering
        \includegraphics[width=0.16\textwidth]{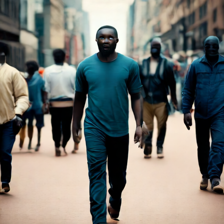}\hfill
        \includegraphics[width=0.16\textwidth]{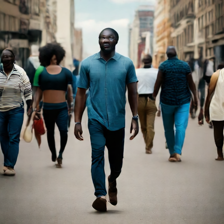}\hfill
        \includegraphics[width=0.16\textwidth]{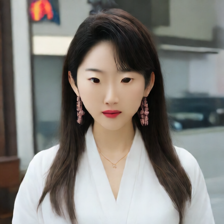}\hfill
        \includegraphics[width=0.16\textwidth]{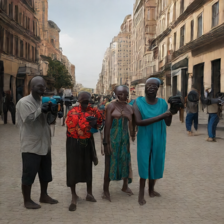}\hfill
        \includegraphics[width=0.16\textwidth]{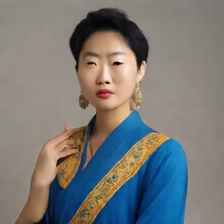}
        \caption{Show-o2 -- ``a photorealistic photo of an african male walking in the city with other people in the background''}
        \label{fig:lang-comparison-show}
    \end{subfigure}
    \caption{Images generated by Janus, MMaDA, and Show-o2 with prompts in English, Spanish, Japanese, Turkish, and Chinese (columns, left to right).}
    \label{fig:lang-comparison}
\end{figure*}

The accuracies for each model are shown in \Cref{tab:language}. As for Bagel, Emu, Gemini, and ChatGPT, the images produced by different languages are inseparable, as the accuracy is the same as a random guess of 20\%. In Janus, we observed that while images generated by English, Spanish, and Chinese are higher-quality, images for Japanese and Turkish prompts are often scenery that is unrelated to the prompts, with most of them being a tower or a mountain. In particular, in many of the images for Japanese prompts and some of the images for Chinese prompts, there is a temple-like house on the bottom left (\Cref{fig:lang-comparison-janus}). As for MMaDA, we observed that the images generated by Japanese and Turkish prompts often do not follow the prompts and are just some abstract objects (\Cref{fig:lang-comparison-mmada}). This explains the high accuracy for Japanese and Turkish prompts and moderately high confusion of Japanese and Turkish prompts (\Cref{tab:mmada-matrix}). We observed that Show-o2 generates a picture of Asian women with high likelihood when the prompt language is Japanese or Chinese, regardless of the prompt semantics (\Cref{fig:lang-comparison-show}).

\begin{table}[t]
    \centering
    \begin{tabular}{l r}
        Model & Accuracy (\%) \\
        \midrule
        BAGEL   & 21.2 \\
        Emu     & 21.9 \\
        Janus   & 52.9 \\
        MMaDA   & 34.2 \\
        Show-o2 & 54.0 \\
        Gemini  & 20.1 \\
        ChatGPT & 21.9 \\
    \end{tabular}
    \caption{\textbf{Most models generate language-agnostic images, but Janus and Show-o2 reveal language-specific visual artifacts.} Accuracy on 5-way prompt language classification from generated images; random chance is 20\%.}
    \label{tab:language}
\end{table}

\begin{table}[t]
\centering
\resizebox{0.9\columnwidth}{!}{
\begin{tabular}{lw{c}{1.2cm}w{c}{1.2cm}w{c}{1.2cm}w{c}{1.2cm}w{c}{1.2cm}}
\diagbox{\textbf{True}}{\textbf{Pred}} & \rotatebox{45}{en} & \rotatebox{45}{es} & \rotatebox{45}{ja} & \rotatebox{45}{tr} & \rotatebox{45}{zh} \\
\text{en} & \cellcolor[rgb]{0.130,0.560,0.552}\parbox[c][1.2cm][c]{1.2cm}{\centering \textcolor{white}{30.0}} & \cellcolor[rgb]{0.267,0.228,0.514}\parbox[c][1.2cm][c]{1.2cm}{\centering \textcolor{white}{10.7}} & \cellcolor[rgb]{0.146,0.515,0.557}\parbox[c][1.2cm][c]{1.2cm}{\centering \textcolor{white}{27.3}} & \cellcolor[rgb]{0.149,0.508,0.557}\parbox[c][1.2cm][c]{1.2cm}{\centering \textcolor{white}{26.7}} & \cellcolor[rgb]{0.283,0.111,0.432}\parbox[c][1.2cm][c]{1.2cm}{\centering \textcolor{white}{5.3}} \\
\text{es} & \cellcolor[rgb]{0.186,0.419,0.557}\parbox[c][1.2cm][c]{1.2cm}{\centering \textcolor{white}{21.3}} & \cellcolor[rgb]{0.211,0.364,0.552}\parbox[c][1.2cm][c]{1.2cm}{\centering \textcolor{white}{18.0}} & \cellcolor[rgb]{0.120,0.604,0.541}\parbox[c][1.2cm][c]{1.2cm}{\centering \textcolor{black}{32.7}} & \cellcolor[rgb]{0.158,0.486,0.558}\parbox[c][1.2cm][c]{1.2cm}{\centering \textcolor{white}{25.3}} & \cellcolor[rgb]{0.277,0.050,0.376}\parbox[c][1.2cm][c]{1.2cm}{\centering \textcolor{white}{2.7}} \\
\text{ja} & \cellcolor[rgb]{0.274,0.200,0.499}\parbox[c][1.2cm][c]{1.2cm}{\centering \textcolor{white}{9.3}} & \cellcolor[rgb]{0.283,0.100,0.422}\parbox[c][1.2cm][c]{1.2cm}{\centering \textcolor{white}{4.7}} & \cellcolor[rgb]{0.993,0.906,0.144}\parbox[c][1.2cm][c]{1.2cm}{\centering \textcolor{black}{60.0}} & \cellcolor[rgb]{0.194,0.399,0.556}\parbox[c][1.2cm][c]{1.2cm}{\centering \textcolor{white}{20.0}} & \cellcolor[rgb]{0.283,0.126,0.445}\parbox[c][1.2cm][c]{1.2cm}{\centering \textcolor{white}{6.0}} \\
\text{tr} & \cellcolor[rgb]{0.283,0.100,0.422}\parbox[c][1.2cm][c]{1.2cm}{\centering \textcolor{white}{4.7}} & \cellcolor[rgb]{0.283,0.126,0.445}\parbox[c][1.2cm][c]{1.2cm}{\centering \textcolor{white}{6.0}} & \cellcolor[rgb]{0.126,0.571,0.550}\parbox[c][1.2cm][c]{1.2cm}{\centering \textcolor{black}{30.7}} & \cellcolor[rgb]{0.916,0.896,0.101}\parbox[c][1.2cm][c]{1.2cm}{\centering \textcolor{black}{58.0}} & \cellcolor[rgb]{0.267,0.005,0.329}\parbox[c][1.2cm][c]{1.2cm}{\centering \textcolor{white}{0.7}} \\
\text{zh} & \cellcolor[rgb]{0.181,0.430,0.557}\parbox[c][1.2cm][c]{1.2cm}{\centering \textcolor{white}{22.0}} & \cellcolor[rgb]{0.277,0.185,0.490}\parbox[c][1.2cm][c]{1.2cm}{\centering \textcolor{white}{8.7}} & \cellcolor[rgb]{0.153,0.497,0.558}\parbox[c][1.2cm][c]{1.2cm}{\centering \textcolor{white}{26.0}} & \cellcolor[rgb]{0.120,0.622,0.535}\parbox[c][1.2cm][c]{1.2cm}{\centering \textcolor{black}{34.0}} & \cellcolor[rgb]{0.274,0.200,0.499}\parbox[c][1.2cm][c]{1.2cm}{\centering \textcolor{white}{9.3}} \\
\end{tabular}
}
\caption{\textbf{MMaDA ignores Japanese and Turkish prompts and generates similar abstract images for both, causing the classifier to frequently confuse the two languages.} True is the ground-truth label; Pred is the classifier prediction.}
\label{tab:mmada-matrix}
\end{table}

\section{Conclusion}

As synthetic imagery increasingly circulates, practical mechanisms for distinguishing and attributing model-generated images are needed to support downstream analysis, monitoring, and measurement of generative ecosystems. Our results show that such attribution is feasible, a classifier trained on images from five open-source unified models substantially outperforms chance, achieving 93.9\% accuracy with only 3K images per model and approaching perfect performance at larger scales. Attribution remains robust under corruptions and structural perturbations, indicating that separability is driven by high-level visual characteristics rather than low-level artifacts or semantic content.

At the same time, our findings reveal clear limits to what can be inferred from generated images alone. While model identity can often be recovered with high confidence, prompt language does not reliably influence the distribution of generated images once the generating model is fixed. Across five languages and both open- and closed-source models, language attribution remains near chance, suggesting that linguistic variation leaves weak visual traces in current unified systems. Together, these results indicate that unified multimodal models exhibit stable, distinguishable generation behavior at the model level, while finer-grained contextual signals remain difficult to recover. This delineates the scope of current image attribution methods and highlights important directions for future work on understanding, characterizing, and monitoring large-scale synthetic image generation.

There are several promising directions for future work. One is to apply mechanistic interpretability techniques to better understand which visual features drive separability, and why certain factors make attribution more difficult. Another avenue is to investigate how prompt complexity influences separability, particularly whether richer prompts reduce the degree to which model-specific biases and artifacts appear in the generated images. Finally, exploring additional sources of variation, such as prompt style, output resolution, and changes in model parameters, may further help characterize what governs separability in unified model-generated images.
\section*{Limitations}

This work studies seven unified models at one point in time; rapidly evolving model architectures and training procedures may shift the separability landscape. Our corruption and perturbation experiments are limited to the MJHQ-30K domain, and the OOD experiment covers ten broad semantic domains that may not represent all real-world use cases. The language experiment covers five languages and may not generalize to other linguistic families. Attribution accuracy also decreases at very small training set sizes, so deployment in low-data regimes requires care.

\section*{Acknowledgments}

We would like to thank Zhuang Liu for his invaluable support and resources throughout this project. 

\bibliography{custom}

\newpage
\onecolumn
\appendix

\section{MLLM Attribution: Prompts and Experimental Setup}
\label{app:mllm_eval}

We evaluate two frontier MLLMs as attributors: GPT-5.4 and Gemini 3.1 Flash. Each evaluator is shown a target image and asked to identify which of the five open-source candidate unified models generated it (BAGEL, Emu3.5, Janus-Pro-7B, MMaDA, Show-o2). We evaluate on 1{,}000 images per source model, sampled from MJHQ-30K prompts, for a total of 5{,}000 queries per evaluator per shot condition.

\paragraph{Zero-shot prompt.} The following prompt is appended after the target image:

\begin{quote}
\small
This image was generated by one of the following AI image generation models:\\
- BAGEL\\
- Emu3.5\\
- Janus-Pro-7B\\
- MMaDA\\
- Show-o2

Based on visual characteristics alone, which model do you think generated this image?

Think step by step about the visual characteristics that distinguish these models (e.g.\ style, artifacts, color palette, sharpness, composition). Then on the very last line of your response, write only the exact model name from the list above, nothing else.
\end{quote}

\paragraph{Few-shot setup.} For $k$-shot evaluation ($k \in \{1, 5\}$), we prepend $k$ example sets before the target question. Each example set contains one image per candidate model drawn from a randomly sampled image ID distinct from the target; the order of models within each set is randomized independently. The example block is prefixed with \textit{``Below are $k$ reference example(s) per model so you can see each model's visual style,''} with each image followed by its ground-truth model name. After the examples, the evaluator is shown the target image with the same chain-of-thought instruction as in the zero-shot case.

\end{document}